\documentclass{article}

\usepackage[T1]{fontenc}
\usepackage{microtype}
\usepackage{graphicx}
\usepackage{booktabs}   
\usepackage{multirow}   
\usepackage{amsmath}
\usepackage{tikz}
\usepackage{xcolor}
\usepackage{xspace}
\usepackage{eso-pic}    
\usepackage[most]{tcolorbox}
\usepackage[font=small,labelfont=bf,skip=2pt]{caption}
\usepackage{stmaryrd}
\newcommand*\circled[1]{\tikz[baseline=(char.base)]{
            \node[shape=circle,fill,inner sep=0.8pt] (char) {\small\textcolor{white}{#1}};}\xspace}

\usepackage{hyperref}

\usepackage[accepted]{mlsys2026}

\newcommand{\name}{AgenticCache\xspace}   

\providecommand{\AtPageUpperRight}[1]{%
  \AtPageUpperLeft{\put(\LenToUnit{\paperwidth},0){#1}}}
\makeatletter
\newcommand{\@acmbadgelist}{}
\newcommand{\acmBadgeR}[2][]{%
  \ifx\@acmbadgelist\@empty
    \def\@acmbadgelist{\href{#1}{\includegraphics[height=1.5cm]{#2}}}%
  \else
    \expandafter\def\expandafter\@acmbadgelist\expandafter{%
      \@acmbadgelist\hspace{2mm}\href{#1}{\includegraphics[height=1.5cm]{#2}}}%
  \fi
}
\AtBeginDocument{%
  \ifx\@acmbadgelist\@empty\else
    \AddToShipoutPictureBG*{%
      \AtPageUpperRight{%
        \raisebox{-2.0cm}{%
          \llap{\@acmbadgelist\hspace{28mm}}%
        }%
      }%
    }%
  \fi
}
\makeatother
\acmBadgeR[https://www.acm.org/publications/policies/artifact-review-and-badging-current]{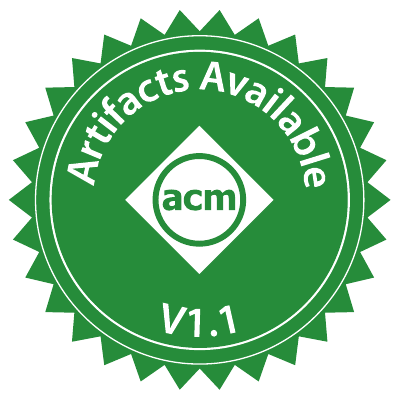}
\acmBadgeR[https://www.acm.org/publications/policies/artifact-review-and-badging-current]{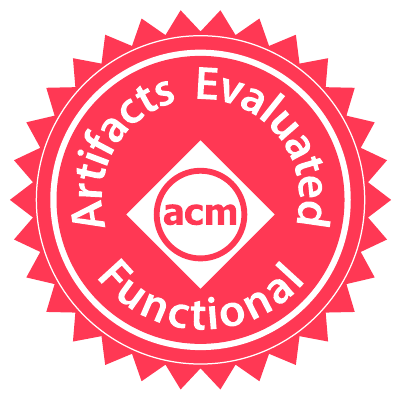}
\acmBadgeR[https://www.acm.org/publications/policies/artifact-review-and-badging-current]{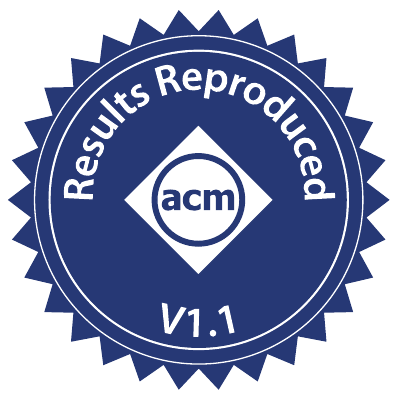}

\mlsystitlerunning{\name: Cache-Driven Asynchronous Planning for Embodied AI Agents}
\begin{document}

\twocolumn[
\mlsystitle{\texorpdfstring{\name: Cache-Driven Asynchronous Planning for Embodied AI Agents}{\name: Cache-Driven Asynchronous Planning for Embodied AI Agents}}



\mlsyssetsymbol{equal}{*}

\begin{mlsysauthorlist}
\mlsysauthor{Hojoon Kim}{snu}
\mlsysauthor{Yuheng Wu}{su}
\mlsysauthor{Thierry Tambe}{su}
\end{mlsysauthorlist}

\mlsysaffiliation{snu}{Seoul National University, Seoul, South Korea}
\mlsysaffiliation{su}{Stanford University, Stanford, California, USA}

\mlsyscorrespondingauthor{Hojoon Kim}{hojoon.kim@snu.ac.kr}
\mlsyscorrespondingauthor{Thierry Tambe}{ttambe@stanford.edu}

\mlsyskeywords{Embodied AI Agents, LLM-based Planning, Plan-Level Locality, Plan Caching, Asynchronous Planning, LLM Inference Efficiency}

\vskip 0.3in

\begin{abstract}
Embodied AI agents increasingly rely on large language models (LLMs) for planning, yet per-step LLM calls impose severe latency and cost. In this paper, we show that embodied tasks exhibit strong plan locality, where the next plan is largely predictable from the current one. Building on this, we introduce \name, a planning framework that reuses cached plans to avoid per-step LLM calls. In \name, each agent queries a runtime cache of frequent plan transitions, while a background Cache Updater asynchronously calls the LLM to validate and refine cached entries. Across four multi-agent embodied benchmarks, \name improves task success rate by 22\% on average across 12 configurations (4 benchmarks $\times$ 3 models), reduces simulation latency by 65\%, and lowers token usage by 50\%. Cache-based plan reuse thus offers a practical path to low-latency, low-cost embodied agents. Code is available at \url{https://github.com/hojoonleokim/MLSys26_AgenticCache}.
\end{abstract}
]



\printAffiliationsAndNotice{}  

\section{Introduction}
\label{sec:introduction}

Embodied AI aims to build agents that perceive, plan, and act to complete tasks in their environments~\cite{zhang2024building,zhang2025combo,liu2025coherent}.
Traditional approaches implement this perceive-plan-act loop through handcrafted, domain-specific pipelines.
While effective in narrow settings, such pipelines demand task engineering and become brittle once the environment shifts~\cite{liu2025aligning}.

Advances in large language models (LLMs) have offered a general and flexible alternative to these manually designed pipelines~\cite{yao2022react,shinn2023reflexion,li2023camel,park2024generative,hong2024metagpt}.
By interpreting perceptual inputs, generating high-level plans, and guiding downstream actions, LLMs enable a unified decision-making framework for embodied agents~\cite{park2023generative,wang2024voyager}, eliminating the need for task engineering.

\begin{figure}[t]
  \centering
  \includegraphics[width=\linewidth]{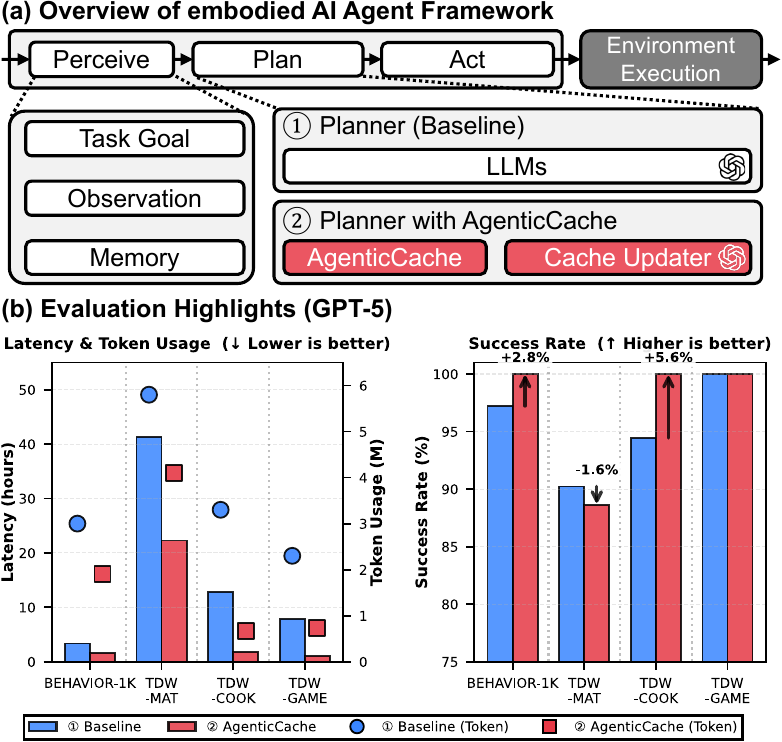}
  \caption{Overview of \name. (a) Embodied AI agent framework. (b) Evaluation highlights on GPT-5.}
  \label{fig:EmbodiedAIAgent}
\end{figure} 

However, invoking LLMs inside this loop introduces significant inference latency and cost. Under the standard synchronous setup, the agent must wait for each plan before acting, stalling real-time execution. To reduce this latency, recent work explores parallelized planning-acting~\cite{li2026parallelized}, which overlaps plan generation with ongoing actions; and speculative planning~\cite{hua2025interactive}, which uses a smaller LLM to propose preliminary actions that a larger LLM later verifies. Yet, these approaches still rely on LLM calls at every step, leaving runtime overhead.

In practice, the next plan is often predictable from the local context, a property we refer to as plan locality~\cite{sutton1998reinforcement}. For example, once an object has been grasped, placing it at the target location is often the natural next step. Humans naturally exploit such regularities through accumulated experience, forming an internal cache of short-horizon plans that enables fast, intuitive responses without deliberate reasoning at every step~\cite{daw2005uncertainty,botvinick2019reinforcement}. Inspired by this observation, we ask: 
\begin{center}
\begin{minipage}{0.92\columnwidth}
\centering
\textit{Can embodied agents similarly leverage a cache-based mechanism to reuse plans and avoid per-step LLM calls, thereby reducing latency and cost?}
\end{minipage}
\end{center}

In this paper, we show that they can, because embodied tasks exhibit strong plan locality. We introduce \name, which reuses cached plans so the agent avoids calling the LLM at every step. As illustrated in Figure~\ref{fig:EmbodiedAIAgent}(a), the agent queries a runtime cache of frequent plan transitions at every planning step, while a background Cache Updater asynchronously calls the LLM to validate and refine cached entries. The savings come from two reasons. First, on a cache hit, the agent can chain multiple cached plans and act continuously, completing several actions before a single background LLM query returns. Second, when the LLM response arrives, the updater either confirms the current plan and waits for it to finish before querying again, or immediately swaps in a correction. As the cache accumulates validated transitions during execution, both latency and cost continue to decrease.
As shown in Figure~\ref{fig:EmbodiedAIAgent}(b), \name achieves up to 86\% latency reduction and 79\% cost savings with GPT-5 on TDW-COOK, while maintaining a 97\% average task success rate across four benchmarks.

\name is complementary to modern LLM serving and efficiency techniques~\cite{kwon2023efficient,zheng2024sglang,park2025decdec} and compatible with multi-agent simulation frameworks~\cite{xie2025ai}. Our contributions are as follows:
\begin{itemize}
\item We identify plan locality as a key property of embodied tasks, showing plan transitions follow predictable short-horizon patterns that enable cache-based planning.
\item We introduce \name, a runtime cache of plan transitions asynchronously maintained by an LLM updater, enabling efficient embodied AI agents.
\item We evaluate \name across four long-horizon multi-agent benchmarks and three model scales, demonstrating on average a 22\% higher task success rate, 50\% lower token usage, and 65\% lower latency.
\end{itemize}
\section{Background}
\label{sec:background}

In this section, we review LLM-powered embodied agents (Section~\ref{sec:background:agents}), parallel planning strategies (Section~\ref{sec:background:parallelism}), and existing cache mechanisms (Section~\ref{sec:background:caching}). Together, these motivate a cache design for embodied planning.

\subsection{Embodied AI Agents Powered by LLMs}
\label{sec:background:agents}

\paragraph{Embodied AI Framework.}
Figure~\ref{fig:EmbodiedAIAgent}(a) illustrates the embodied AI framework. 
An embodied agent typically operates through three core stages: perceive, plan, and act. 
It first perceives the environment by gathering observations, tracking task goals, and maintaining memory. 
It then plans by decomposing long-horizon objectives into subgoals. 
Finally, it acts by executing these actions in the environment.
The environment is then updated, yielding new observations for the next round of perception and planning.

\begin{figure}[t]
  \centering
  \includegraphics[width=\linewidth]{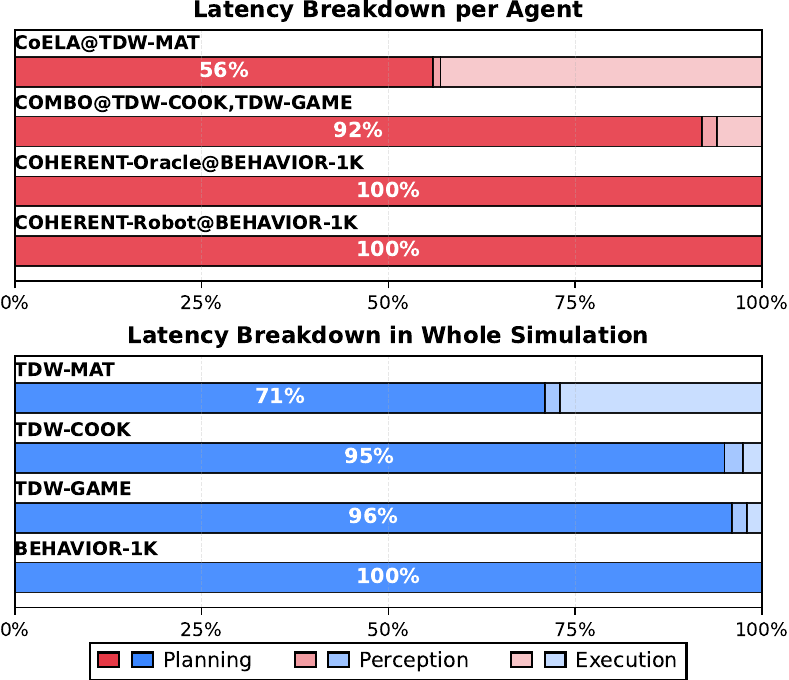}
  \caption{Latency breakdown across agents and benchmarks.}
  \label{fig:baseline_breakdown}
\end{figure}

\paragraph{LLM-Powered Embodied Agents.}
Recent advances in training and inference have significantly improved the reasoning capabilities of LLMs~\cite{guo2025deepseek,brown_2024_large,wu2025on,wu2025tom}. As a result, embodied AI systems increasingly use LLMs as reasoning cores that process perceptual inputs and generate high-level plans~\cite{yao2022react,shinn2023reflexion,li2023camel}. However, this design creates substantial latency and cost bottlenecks. As shown in Figure~\ref{fig:baseline_breakdown}, over 70\% of runtime across benchmarks is spent on LLM queries for planning. At each step, the agent must wait for the LLM's response before acting, forming a synchronous plan-act loop as illustrated in Figure~\ref{fig:comparison}(a).

\subsection{Planning Parallelism in Embodied AI Agents}
\label{sec:background:parallelism}
\begin{figure}[t]
  \centering
  \includegraphics[width=\linewidth]{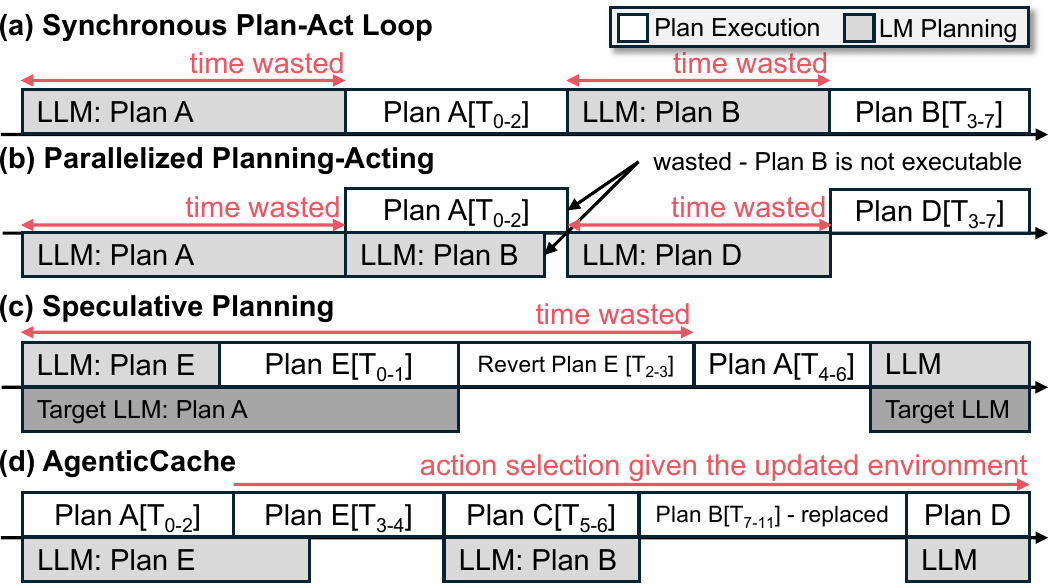}
  \caption{Comparison of four planning strategies. (a) Synchronous plan-act loop. (b) Parallelized planning-acting. (c) Speculative planning. (d) \name.}
  \label{fig:comparison}
\end{figure}

\paragraph{Asynchronous Parallel Planning Strategies.}
To address the latency bottleneck of synchronous planning, recent work explores asynchronous planning strategies:

(1) \textit{Parallelized Planning-Acting}~\cite{li2026parallelized}. Queries the next plan while executing the current one, partially hiding LLM latency. However, as shown in Figure~\ref{fig:comparison}(b), the generated plan may become invalid when the environment changes, requiring replanning and adding runtime overhead.

(2) \textit{Speculative Planning}~\cite{hua2025interactive}. Uses a smaller LLM to propose actions that are later verified by a larger LLM. However, as shown in Figure~\ref{fig:comparison}(c), it performs poorly in realistic environments where correcting wrong actions, such as moving back or undoing manipulations, takes time.

\paragraph{Limitations of Existing Parallel Planning.} 
A limitation of both methods is their reliance on repeated LLM queries for plan generation, so the LLM cost scales linearly with the trajectory length. Neither method exploits the recurring structure of plans across timesteps, treating every step as a fresh decision even when the current and next plans follow a predictable pattern. Reusing familiar plan patterns through a cache, as discussed in Section~\ref{sec:introduction}, offers an alternative that avoids per-step LLM queries, which we explore next.

\subsection{Caching Mechanisms}
\label{sec:background:caching}

Caching is a common way to reduce redundant computation in LLM systems. Existing approaches cache reusable information at several levels, from token-level activations to full responses and higher-level task patterns:

(1) \textit{KV Cache}~\cite{kwon2023efficient,zheng2024sglang}. Stores key-value states from previous decoding steps, allowing the model to reuse them when generating subsequent tokens without recomputing the entire prefix.

(2) \textit{Context Cache}~\cite{bang2023gptcache,hu2024memserve,hu2025epic}. Caches prompt-response pairs or similar contexts. When a new request matches a cached entry, the system can reuse the cached response instead of invoking the model again.

(3) \textit{Template Cache}~\cite{zhang2025cost,ruan2025cortex}. Caches plan templates or structured outputs associated with recurring task patterns. Cache hits are determined by similarity or pattern matching against stored templates.

\paragraph{Limitations of Existing Caches.}
Existing caching mechanisms are primarily designed for LLM inference rather than embodied agents. They reduce redundant computation inside the model, but do not reduce repeated LLM invocations during execution. In the next section, we show that embodied tasks exhibit strong plan locality, which \name directly exploits for a cache-based planner.

\section{Plan Locality in Embodied AI Agents}
\label{sec:accelerate}

\begin{figure}[t]
  \centering
  \includegraphics[width=\linewidth]{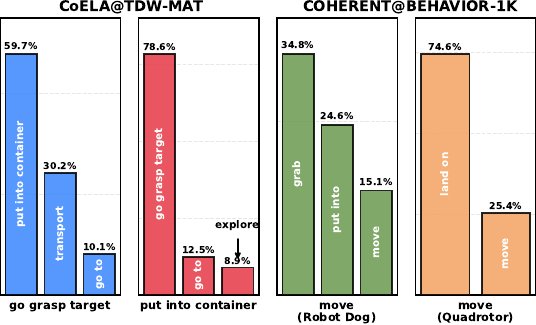}
\caption{Probability distribution of successor plans under a 2-gram model from GPT-5 execution trajectories.}
  \label{fig:ActionLocality}
\end{figure}

In this section, we show that embodied AI tasks exhibit widespread plan locality (Section~\ref{sec:accelerate:locality}), yet locality alone is insufficient under dynamic environments, motivating \name's hybrid cache-LLM design (Section~\ref{sec:accelerate:context}).

\subsection{Plan Locality}
\label{sec:accelerate:locality}

\paragraph{2-Gram Plan Locality.}
Long-horizon embodied tasks exhibit strong plan locality, where certain plan transitions occur with high regularity. Figure~\ref{fig:ActionLocality} presents a 2-gram analysis of plan transitions, showing that many plans have only a small set of likely successors. For example, after executing ``go grasp target,'' the next plan is ``put into container'' in 59.7\% of cases. This regularity suggests that embodied agents often follow stable short-horizon patterns rather than switching arbitrarily between plans, making cached reuse of familiar plans a promising strategy.

\subsection{Beyond Pure Locality}
\label{sec:accelerate:context}

\begin{figure}[t]
  \centering
    \includegraphics[width=\linewidth]{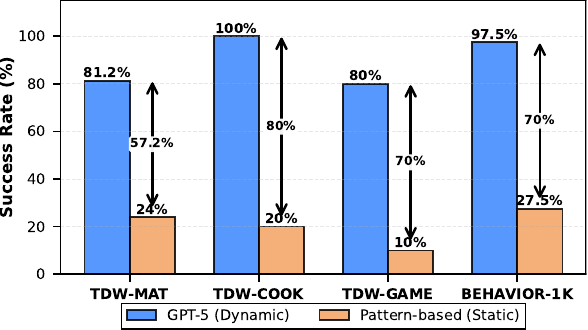}
  \caption{Pattern-based agents exploit plan locality but suffer large performance gaps without context-aware updates.}
  \label{fig:cache_vs_dynamic}
\end{figure}

\paragraph{Limitations of Pure Locality.}
As shown in Figure~\ref{fig:cache_vs_dynamic}, simply following cached plan patterns without considering the evolving environment leads to performance degradation compared to GPT-5 agents. This shows that plan locality alone is insufficient: while many transitions are predictable, environmental changes can invalidate cached plans. For example, a cached \texttt{GoGrasp} transition may fail if another agent has already picked up the target object, or the environment has moved it out of reach.

\paragraph{Toward Hybrid Planning.}
To remain robust in such settings, agents must combine fast cache-based reuse with selective LLM reasoning. Pure cache reuse risks acting on stale information, while always consulting the LLM reintroduces the latency that caching is meant to avoid. This hybrid mechanism allows the agent to act efficiently in familiar contexts while invoking the planner only when new or uncertain situations arise. This need for both efficiency and contextual adaptability motivates the design of \name.

\section{\name Design}
\label{sec:agenticcache}

In this section, we present \name's design, covering the cache as a local planner (Section~\ref{sec:cache-system}), the asynchronous Cache Updater (Section~\ref{sec:updater-process}), an optional warm-start strategy (Section~\ref{sec:cold-miss}), and a runtime example (Section~\ref{sec:runtime-example}).

\subsection{\name as a Planner}
\label{sec:cache-system}

\begin{figure*}[t]
  \centering
  \includegraphics[width=\linewidth]{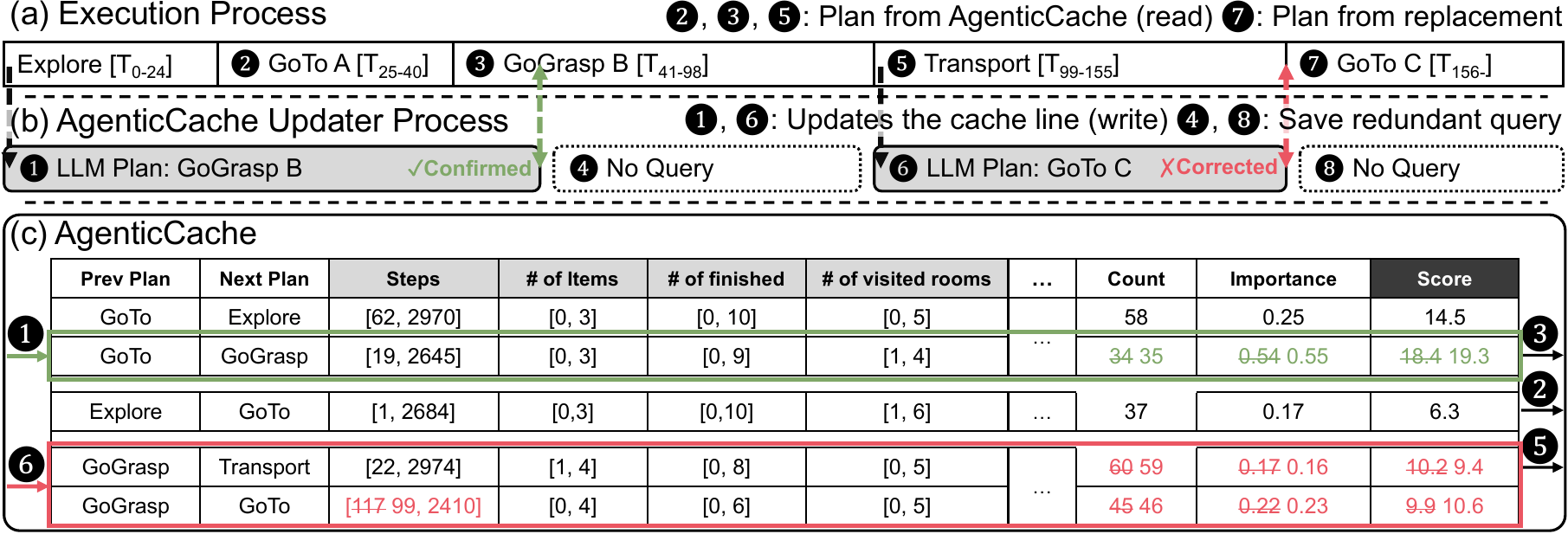}
  \caption{Runtime example of \name execution.}
  \label{fig:AgenticCache_Example}
\end{figure*}

\paragraph{\name Structure.}
Each embodied agent maintains its own cache, which serves as a lightweight planner storing frequent plan-to-plan transitions. 
As illustrated in Figure~\ref{fig:AgenticCache_Example}(c), each cache entry is represented as a 2-gram pattern $\langle P_i \!\rightarrow\! P_j\rangle$, 
where $P_i$ and $P_j$ denote consecutive high-level plans 
(e.g., \texttt{GoGrasp} $\!\rightarrow$ \texttt{Transport}). 
In addition to transition statistics, each entry records a set of task-state metadata fields extracted from both offline and online episodes.
The specific metadata fields depend on the information available to each agent and the benchmark environment.
For each field, the cache stores the observed minimum and maximum as an integer range, 
capturing the state conditions under which the transition has historically occurred.
Figure~\ref{fig:AgenticCache_Example} shows a concrete example from a TDW-MAT task, 
where the metadata includes the episode step index (\textbf{Steps}), the number of held objects (\textbf{\# of Items}), 
the number of completed sub-goals (\textbf{\# of finished}), and the number of visited rooms (\textbf{\# of visited rooms}).
Other benchmarks use a different subset of state features, typically object counts and progress toward sub-goals.

\paragraph{Runtime Query and Filtering.}
During execution, the control loop queries the cache with the previous plan $P_i$ and the current task-state metadata as keys. A filtering stage removes entries whose metadata ranges conflict with the current state, yielding the feasible candidate set:
\begin{equation*}
\mathcal{F}(P_i) = \{\, P_j \mid s_t \in \llbracket s^{\min}_{ij}, s^{\max}_{ij} \rrbracket,\ 
h_t \in \llbracket h^{\min}_{ij}, h^{\max}_{ij} \rrbracket \,\}.
\end{equation*}

\paragraph{Scoring and Selection.}
The next plan is selected by maximizing a composite score among feasible candidates:
\begin{equation*}
P^* = \arg\max_{P_j \in \mathcal{F}(P_i)} S(P_i \!\rightarrow\! P_j),
\end{equation*}
where the score is defined as
\begin{equation*}
S(P_i \!\rightarrow\! P_j) = C(P_i \!\rightarrow\! P_j) \cdot I(P_j).
\end{equation*}
The two factors capture complementary signals. $C(P_i \!\rightarrow\! P_j)$ is the \textbf{transition count}, i.e., how many times plan $P_j$ has been observed immediately after $P_i$ during execution. A high $C$ means the transition is frequent, but frequency alone can be misleading, since a locally frequent transition may not be globally reliable. The \textbf{importance factor} $I(P_j)$ compensates for this by measuring how often $P_j$ is confirmed by the background LLM:
\begin{equation*}
I(P_j) = \frac{N^{\text{conf}}(P_j)}{N^{\text{cand}}(P_j)},
\end{equation*}
where $N^{\text{cand}}(P_j)$ is the number of times $P_j$ has appeared as a feasible candidate during cache queries, and $N^{\text{conf}}(P_j)$ is the number of times the background LLM subsequently confirmed $P_j$ as the correct plan.

This design mirrors hybrid branch predictors that combine local and global history~\cite{yeh1993comparison,smith1998study}: $C$ is a \emph{local} signal specific to the $P_i \!\rightarrow\! P_j$ transition, while $I$ is a \emph{global} signal aggregated across all contexts in which $P_j$ appeared as a candidate. Their product rewards transitions that are both locally frequent and globally reliable, so neither signal alone dominates. In Figure~\ref{fig:AgenticCache_Example}(c), the \textbf{Count}, \textbf{Importance}, and \textbf{Score} columns correspond to $C(P_i \!\to\! P_j)$, $I(P_j)$, and $S(P_i \!\to\! P_j)$, respectively.

\subsection{\name Updater}
\label{sec:updater-process}
The \name Updater (Figure~\ref{fig:AgenticCache_Example}(b)) is a background LLM process that maintains cache quality during execution. It asynchronously queries the LLM to validate, correct, and refine cache entries without blocking execution.

\paragraph{Update Mechanism.}
When issuing a query, the updater records the current context $c_t$ and the active plan $p_t$. 
After an asynchronous delay of $k$ steps, it receives the LLM's response $p'_{t+k}$ and compares this prediction against the executed plan trajectory $\{p_{t+1}, \ldots, p_{t+k}\}$ to determine whether the cache's choice was correct.

(1) \textit{Confirmation.}  
If $p'_{t+k}$ already appears in the executed trajectory, the cache has correctly anticipated the LLM's choice. 
The updater reinforces the corresponding transition by incrementing its transition count $C(p_t \!\rightarrow\! p'_{t+k})$ and confirmation count $N^{\text{conf}}(p'_{t+k})$.

(2) \textit{Correction.}  
If $p'_{t+k}$ does not appear in the recent trajectory, the cache mispredicted the LLM's preferred plan. 
The updater then (a) adds a new or updated transition for $p_t \!\rightarrow\! p'_{t+k}$,  
(b) decreases the counts of the mispredicted transition, and  
(c) replaces the ongoing plan with $p'_{t+k}$ if it is executable. 
This immediate replacement preserves robustness under stale cache hits. Rather than waiting for the current cached plan to fully terminate, the agent switches to the corrected plan as soon as the updater detects that the cached plan is no longer appropriate.

\paragraph{Query Control.}
The updater issues LLM queries periodically throughout execution, but suppresses redundant queries in two cases described below.

\textit{(1) Confirmation suppression.} When the LLM response confirms the currently executing plan, no further queries are issued until that plan finishes. Since the cache and LLM agree on the current action, queries during this window would yield no new information and consume tokens.

\textit{(2) Correction suppression.} After a correction replaces the ongoing plan, additional queries are withheld until the replacement plan completes execution, preventing conflicting corrections that could destabilize ongoing execution.

\paragraph{Cache Miss.}
A cache miss occurs when no feasible plan can be retrieved, either because the situation is unseen or all candidates are filtered out by the metadata ranges.
High-level planning temporarily pauses, and the updater requests a new plan from the LLM. 
The new plan is inserted as a fresh cache entry, and the agent resumes execution.

\paragraph{Summary.}
Together, the cache and updater form a symbiotic system. The cache functions as a fast, pattern-based planner that exploits historical regularities, while the updater provides context-aware correction grounded in LLM reasoning.
This asynchronous cooperation enables continuous, low-latency decision-making while preserving adaptability in dynamic environments.

\subsection{Offline Pattern Prefilling for Warm-Start}
\label{sec:cold-miss}

\paragraph{Motivation.}
Like caches in computer architecture, \name suffers cold misses when an episode starts with no prior transitions. The agent must then wait for the LLM to produce the first few updates, reintroducing the latency that caching aims to avoid. To mitigate this, we provide an optional offline pattern prefilling procedure.

Before execution, we initialize the cache with plan-to-plan transitions extracted from successful GPT-5 trajectories on out-of-distribution tasks. Each transition is inserted with its estimated conditional probability and metadata range, producing a warm-start cache.

\paragraph{Role of Prefilling.}
Prefilling lets the agent act immediately from the beginning of an episode, avoiding cold-start delay while the cache continues to evolve online. However, it is not a prerequisite. The updater bootstraps an empty cache online just as well, and our cold-start evaluation (Section~\ref{sec:evaluation}) confirms that \name retains most of its latency and cost savings even without prefilling. In practice, prefilling mainly improves the first several decisions of an episode, while the core benefits of asynchronous cache-guided planning emerge once the cache is populated online.

\subsection{Runtime Workflow Example of \name}
\label{sec:runtime-example}

Figure~\ref{fig:AgenticCache_Example} walks through how \name and its Updater interact throughout an episode.

\paragraph{Cache-Guided Planning.}
At the start of the episode, the agent executes an initial \texttt{Explore} plan while the updater issues a periodic LLM query on the current observation metadata (Step~\circled{1} in Figure~\ref{fig:AgenticCache_Example}(b)). When \texttt{Explore} finishes, the agent queries the cache with it as the previous plan; since no entries are filtered out, the cache returns \texttt{GoTo~A} as the highest-scoring candidate (Step~\circled{2} in Figure~\ref{fig:AgenticCache_Example}(a)). After \texttt{GoTo~A} completes at step 40, the cache is queried again; entries such as \texttt{GoTo}~$\!\rightarrow\!$~\texttt{Explore} are filtered out by step-range mismatch, and \texttt{GoGrasp~B} is selected (Step~\circled{3}).

\paragraph{Asynchronous LLM Feedback.}
During the execution of \texttt{GoGrasp~B}, the earlier LLM query (from Step~\circled{1}) returns, confirming \texttt{GoGrasp~B} as the correct next plan. Since this matches the currently executing plan, the system increments the transition's count and importance (\circled{1} in Figure~\ref{fig:AgenticCache_Example}(c)) and applies confirmation suppression, withholding further queries until \texttt{GoGrasp~B} completes (Step~\circled{4}).

\paragraph{Cache Correction.}
After completing \texttt{GoGrasp~B}, the agent queries the cache again with updated metadata. Transitions such as \texttt{GoGrasp}~$\!\rightarrow\!$~\texttt{GoTo} are filtered out, while \texttt{GoGrasp}~$\!\rightarrow\!$~\texttt{Transport} achieves the highest score and is executed (Step~\circled{5}). Since a new plan has begun and no suppression is active, the updater issues its next periodic LLM query (Step~\circled{6}). While executing \texttt{Transport}, the LLM response arrives, proposing \texttt{GoTo~C} as the next plan. Since \texttt{GoTo~C} does not appear in the recent trajectory, the system replaces the ongoing plan with \texttt{GoTo~C} (Step~\circled{7}). The cache decreases the count and importance of the incorrect transition \texttt{GoGrasp}~$\!\rightarrow\!$~\texttt{Transport} and increments those of the newly proposed transition. The entry's metadata (\texttt{Steps}) is updated from 117 to 99. The updater then applies correction suppression and withholds queries until \texttt{GoTo~C} completes (Step~\circled{8}).

\section{Evaluation}
\label{sec:evaluation}

In this section, we evaluate \name on four multi-agent embodied benchmarks across three model scales (Sections~\ref{sec:evaluationsetup}--\ref{sec:baselines}). We report main results and analyses (Sections~\ref{sec:main-results}--\ref{sec:memory}), ablation and cache validity studies (Sections~\ref{sec:ablation}--\ref{sec:validity}), and close with a discussion (Section~\ref{sec:eval-discussion}). See Appendix~\ref{sec:artifact} for artifact and reproduction details.

\subsection{Evaluation Setup}
\label{sec:evaluationsetup}

\paragraph{Platform and Models.}
All experiments are conducted on a workstation with an NVIDIA GeForce RTX~4090 GPU and an AMD Ryzen~9~7950X 16-core CPU. We use GPT-5, GPT-5-mini, and GPT-5-nano~\cite{openai2025gpt5,openai2025gpt5mini,openai2025gpt5nano} as planners through the OpenAI API~\cite{openai2025api}. Reported cost figures are derived from measured input and output token counts multiplied by OpenAI's listed per-token prices for each model at the time of evaluation (October 2025).

\begin{table}[t]
\centering
\caption{Benchmark characteristics. Plan.: planner modality (LLM or VLM); \#Ag.: number of agents; Coord.: coordination style (Decent.\ = decentralized, Cent.\ = centralized); Env.: simulation environment (TDW = ThreeDWorld, Graph = graph-structured).}
\resizebox{\columnwidth}{!}{%
\begin{tabular}{@{}lccll@{}}
\toprule
Task & Plan. & \#Ag. & Coord. & Env. \\
\midrule
CoELA@TDW-MAT       & LLM & 2 & Decent. & TDW \\
COMBO@TDW-COOK      & VLM & 2 & Decent. & TDW \\
COMBO@TDW-GAME      & VLM & 4 & Decent. & TDW \\
COHERENT@BHV-1K     & LLM & 5 & Cent.   & Graph \\
\bottomrule
\end{tabular}%
}
\label{tab:task_summary}
\end{table}

\paragraph{Cache Prefilling.}
Each benchmark prefills the cache from training episodes disjoint from the evaluation set. We use 4, 2, 1, and 4 training episodes for TDW-MAT, TDW-COOK, TDW-GAME, and BEHAVIOR-1K, respectively, with evaluation on 44, 18, 9, and 36 episodes. The main results (denoted Ours+) use this warm-start configuration; cold-start results without prefilling can be found in Section~\ref{sec:coldstart}.

\begin{table*}[t]
\centering
\footnotesize
\setlength{\tabcolsep}{3pt}
\renewcommand{\arraystretch}{1.1}
\caption{Planning strategy performance across four benchmarks and three model scales. SR: success rate; L: latency (hours); T: token usage; C: cost (USD). Ours+ denotes \name with warm-start cache prefilling.}

{\resizebox{\textwidth}{!}{%
\begin{tabular}{llcccclcccclcccclcccc}
\toprule
\multicolumn{21}{c}{GPT-5} \\ \hline
\multicolumn{1}{c}{\multirow{2}{*}{Execution Strategy}} & \textbf{} & \multicolumn{4}{c}{TDW-MAT} & \textbf{} & \multicolumn{4}{c}{TDW-COOK} & \textbf{} & \multicolumn{4}{c}{TDW-GAME} & \textbf{} & \multicolumn{4}{c}{BEHAVIOR-1K} \\ \cline{3-6} \cline{8-11} \cline{13-16} \cline{18-21} 
\multicolumn{1}{c}{} & \textbf{} & SR & L & T & C & \textbf{} & SR & L & T & C & \textbf{} & SR & L & T & C & \textbf{} & SR & L & T & C \\ \hline
Baseline &  & 90.23\% & 41.34 & 5.8M & \$40.5 &  & 94.44\% & 12.86 & 3.3M & \$21.0 &  & 100\% & 7.88 & 2.3M & \$14.3 &  & 97.22\% & 3.36 & 3.0M & \$9.3 \\
Parallel &  & 89.32\% & 43.67 & 10.4M & \$71.9 &  & 100\% & 14.72 & 6.2M & \$47.6 &  & 0\% & 15.83 & 10.8M & \$84.2 &  & 97.22\% & 3.00 & 4.6M & \$15.3 \\
Speculative &  & 80.91\% & 36.37 & 13.5M & \$39.9 &  & 83.33\% & 6.12 & 7.8M & \$21.6 &  & 11.11\% & 6.13 & 9.9M & \$28.5 &  & 94.44\% & 3.76 & 4.4M & \$10.4 \\
\name (Ours+) &  & 88.64\% & 22.27 & 4.1M & \$27.7 &  & 100\% & 1.75 & 675K & \$4.4 &  & 100\% & 1.11 & 728K & \$4.8 &  & 100\% & 1.55 & 1.9M & \$6.6 \\
\hline
\multicolumn{21}{c}{GPT-5-mini} \\ \hline
\multicolumn{1}{c}{\multirow{2}{*}{Execution Strategy}} & \textbf{} & \multicolumn{4}{c}{TDW-MAT} & \textbf{} & \multicolumn{4}{c}{TDW-COOK} & \textbf{} & \multicolumn{4}{c}{TDW-GAME} & \textbf{} & \multicolumn{4}{c}{BEHAVIOR-1K} \\ \cline{3-6} \cline{8-11} \cline{13-16} \cline{18-21} 
\multicolumn{1}{c}{} & \textbf{} & SR & L & T & C & \textbf{} & SR & L & T & C & \textbf{} & SR & L & T & C & \textbf{} & SR & L & T & C \\ \hline
Baseline &  & 85.45\% & 29.60 & 4.5M & \$5.4 &  & 83.33\% & 6.88 & 2.8M & \$3.2 &  & 22.22\% & 11.06 & 3.9M & \$4.5 &  & 94.44\% & 1.81 & 3.2M & \$1.8 \\
Parallel &  & 84.55\% & 30.84 & 7.5M & \$8.9 &  & 94.44\% & 6.50 & 4.9M & \$6.5 &  & 22.22\% & 6.08 & 6.0M & \$7.5 &  & 91.67\% & 1.63 & 5.0M & \$3.1 \\
Speculative &  & 77.73\% & 37.61 & 12.7M & \$7.7 &  & 50\% & 5.42 & 7.3M & \$4.4 &  & 33.33\% & 4.61 & 8.0M & \$4.7 &  & 94.44\% & 3.05 & 4.5M & \$2.1 \\
\name (Ours+) &  & 84.32\% & 22.56 & 3.6M & \$4.2 &  & 100\% & 1.40 & 826K & \$1.0 &  & 100\% & 1.13 & 840K & \$1.0 &  & 100\% & 0.97 & 1.8M & \$1.2 \\
\hline
\multicolumn{21}{c}{GPT-5-nano} \\ \hline
\multicolumn{1}{c}{\multirow{2}{*}{Execution Strategy}} & \textbf{} & \multicolumn{4}{c}{TDW-MAT} & \textbf{} & \multicolumn{4}{c}{TDW-COOK} & \textbf{} & \multicolumn{4}{c}{TDW-GAME} & \textbf{} & \multicolumn{4}{c}{BEHAVIOR-1K} \\ \cline{3-6} \cline{8-11} \cline{13-16} \cline{18-21} 
\multicolumn{1}{c}{} & \textbf{} & SR & L & T & C & \textbf{} & SR & L & T & C & \textbf{} & SR & L & T & C & \textbf{} & SR & L & T & C \\ \hline
Baseline &  & 71.59\% & 40.20 & 8.8M & \$2.9 &  & 61.11\% & 12.05 & 6.8M & \$2.1 &  & 0\% & 10.07 & 7.8M & \$2.4 &  & 25\% & 10.46 & 9.7M & \$2.4 \\
Parallel &  & 71.59\% & 38.07 & 15.3M & \$5.0 &  & 55.56\% & 9.96 & 13.7M & \$4.4 &  & 0\% & 6.48 & 12.4M & \$3.9 &  & 25\% & 9.27 & 12.8M & \$3.2 \\
\name (Ours+) &  & 67.95\% & 24.62 & 7.6M & \$2.5 &  & 72.22\% & 2.76 & 2.0M & \$0.7 &  & 100\% & 1.08 & 1.2M & \$0.4 &  & 77.78\% & 4.51 & 8.5M & \$2.0 \\
\bottomrule
\end{tabular}
}}
\label{tab:planning_results}
\end{table*}

\begin{figure}
    \centering
    \includegraphics[width=\linewidth]{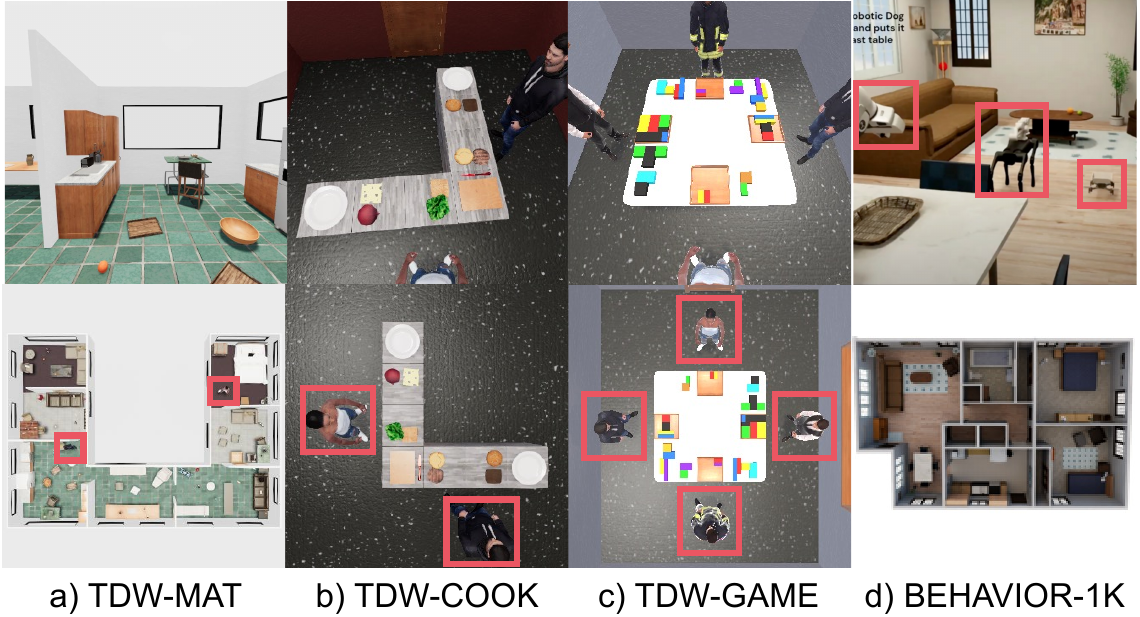}
    \caption{Snapshots from the four benchmark environments, with agents highlighted in red.}
    \label{fig:evaluation:task}
\end{figure}

\paragraph{Benchmarks.}
Table~\ref{tab:task_summary} and Figure~\ref{fig:evaluation:task} summarize the four benchmarks. The three TDW tasks are built on ThreeDWorld~\cite{gan2021threedworld}, a Unity-based 3D simulator.

(1) \textit{TDW-MAT}~\cite{zhang2024building}. A transport task where agents coordinate to move large objects via containers, combining navigation, manipulation, and communication.

(2) \textit{TDW-COOK}~\cite{zhang2025combo}. A cooperative cooking task with agents that follow recipes with strong temporal dependencies across subtasks.

(3) \textit{TDW-GAME}~\cite{zhang2025combo}. A puzzle assembly task requiring VLM reasoning and multi-stage coordination.

(4) \textit{BEHAVIOR-1K}~\cite{li2022behavior1k,li2024behavior1k}. Evaluated with the COHERENT~\cite{liu2025coherent} agent framework, which re-expresses the original simulation as a graph-structured environment and models collaboration among heterogeneous robots (arm, dog, and quadrotor) on household transport tasks. For this benchmark, we merge per-agent LLM queries into a single reasoning call per timestep to reduce communication overhead.

Together, these benchmarks provide a diverse testbed for evaluating how \name improves efficiency while maintaining task performance. All prompts used for these experiments are provided in Appendix~\ref{sec:appendix:prompt}.

\subsection{Baselines}
\label{sec:baselines}

We compare \name against three planning methods:

(1) \textit{Synchronous Baseline.}
We adopt CoELA~\cite{zhang2024building}, COMBO~\cite{zhang2025combo}, and COHERENT~\cite{liu2025coherent} as synchronous LLM-based agents. For CoELA, we apply the planning-then-communication strategy from ReCA~\cite{wan2025reca} to reduce cost and runtime. COMBO reproduction requires training three VLMs and three diffusion models; we use a streamlined variant that preserves its modular plan-act structure, replacing the beam-search process (14 VLM calls and 22 diffusion inferences) with a single VLM and diffusion call. After prompt optimization, our GPT-5 reimplementation achieves success rates comparable to prior reports.

(2) \textit{Parallelized Planning-Acting}~\cite{li2026parallelized}.
Originally proposed for the Odyssey benchmark~\cite{liu2025odyssey}, this approach overlaps planning with execution by issuing LLM queries during ongoing actions. We adapt it to TDW and BEHAVIOR environments. Plans with higher predicted importance preempt current actions, following the priority mechanism from the original paper.

(3) \textit{Speculative Planning}~\cite{hua2025interactive}.
Developed for OpenAGI~\cite{ge2023openagi} and TravelPlanner~\cite{xie2024travelplanner}, this framework executes provisional actions while awaiting high-confidence plans from a stronger model. In our adaptation, GPT-5-nano serves as the lightweight drafter, with GPT-5 or GPT-5-mini as the target model, and we cap speculative depth at three steps.

\subsection{Quantitative Results}
\label{sec:main-results}

\paragraph{Main Results.}
As shown in Table~\ref{tab:planning_results}, \name consistently achieves high task success rates. With GPT-5 and GPT-5-mini it reaches 84--100\% across environments, and with GPT-5-nano it reaches 68--100\%. Baselines struggle in multi-agent settings. On TDW-GAME, parallelized planning achieves 0--22\% and speculative planning 11--33\%, while \name reaches 100\% across all three models. On TDW-COOK, speculative planning drops to 50\% with GPT-5-mini, whereas \name maintains 100\%. Parallelized planning does slightly better on other tasks but still suffers from stale prefetches that become invalid once the environment changes. In contrast, \name's asynchronous updates keep cached plans fresh, avoiding rollbacks and preserving coordinated actions across agents.

\paragraph{Latency and Cost Efficiency.}
\name also substantially reduces latency and cost. Across all 12 configurations, it lowers average latency by 65\% and token consumption by 50\% (Table~\ref{tab:planning_results}). For instance, on TDW-COOK with GPT-5, latency drops from 12.86\,hours to 1.75\,hours (7.4$\times$) and cost from \$21.0 to \$4.4 (4.8$\times$). These gains come from asynchronous cache updates that cut idle waiting and redundant queries, yielding low-latency, low-cost execution.

\subsection{Cold-Start Evaluation Without Offline Prefilling}
\label{sec:coldstart}

To test whether offline prefilling is necessary, we evaluate \name from an empty cache against the synchronous baseline on both standard and long-horizon tasks. Tables~\ref{tab:cold_start_standard} and~\ref{tab:cold_start_long} report results, with ``Ours'' denoting \name with no prefilling.

\begin{table}[t]
\centering
\footnotesize
\setlength{\tabcolsep}{4pt}
\renewcommand{\arraystretch}{1.05}
\caption{Cold-start results on standard tasks. SR: success rate; L: latency (hours); T: token usage; C: cost (USD).}
\label{tab:cold_start_standard}
{
\resizebox{\columnwidth}{!}{
\begin{tabular}{llcccc}
\toprule
Model & Method & SR & L & T & C \\
\hline
GPT-5 & Baseline (CoELA) & 90.0\% & 5.06 & 747K & \$5.32 \\
 & Ours & 93.3\% & 2.63 & 571K & \$3.94 \\
GPT-5-mini & Baseline (CoELA) & 85.0\% & 4.11 & 590K & \$0.75 \\
 & Ours & 85.0\% & 2.67 & 473K & \$0.55 \\
GPT-5-nano & Baseline (CoELA) & 61.7\% & 4.77 & 1.50M & \$0.49 \\
 & Ours & 58.3\% & 3.26 & 1.09M & \$0.36 \\
\bottomrule
\end{tabular}}}
\end{table}

\begin{table}[t]
\centering
\footnotesize
\setlength{\tabcolsep}{4pt}
\renewcommand{\arraystretch}{1.05}
\caption{Cold-start results on long-horizon tasks. SR: success rate; L: latency (hours); T: token usage; C: cost (USD).}
\label{tab:cold_start_long}
{
\resizebox{\columnwidth}{!}{
\begin{tabular}{llcccc}
\toprule
Model & Method & SR & L & T & C \\
\hline
GPT-5 & Baseline (CoELA) & 82.2\% & 11.57 & 1.88M & \$13.09 \\
 & Ours & 80.6\% & 6.19 & 1.08M & \$7.19 \\
GPT-5-mini & Baseline (CoELA) & 62.8\% & 8.17 & 1.25M & \$1.42 \\
 & Ours & 69.4\% & 6.29 & 937K & \$1.04 \\
GPT-5-nano & Baseline (CoELA) & 42.8\% & 11.26 & 2.94M & \$0.92 \\
 & Ours & 62.8\% & 6.63 & 1.94M & \$0.60 \\
\bottomrule
\end{tabular}}}
\end{table}

\paragraph{Results.}
On standard tasks, \name reduces latency by 1.4--1.9$\times$ and cost by 1.35$\times$ across all three model scales while maintaining comparable success rates. On long-horizon tasks, the same efficiency gains persist (1.3--1.9$\times$ latency, 1.4--1.8$\times$ cost), and \name improves success rate with GPT-5-nano (42.8\%$\to$62.8\%) and GPT-5-mini (62.8\%$\to$69.4\%), while slightly trading off for GPT-5 (82.2\%$\to$80.6\%). These results indicate that cold misses cause only transient stall overhead during early cache construction rather than preventing cache-guided planning from being effective. The GPT-5 SR drop likely arises from two effects that long-horizon episodes amplify. First, stale transitions remain locally plausible but become suboptimal after delayed environment changes. Second, independently reused transitions can lead to coordination conflicts.

\begin{figure}
    \centering
    \includegraphics[width=\linewidth]{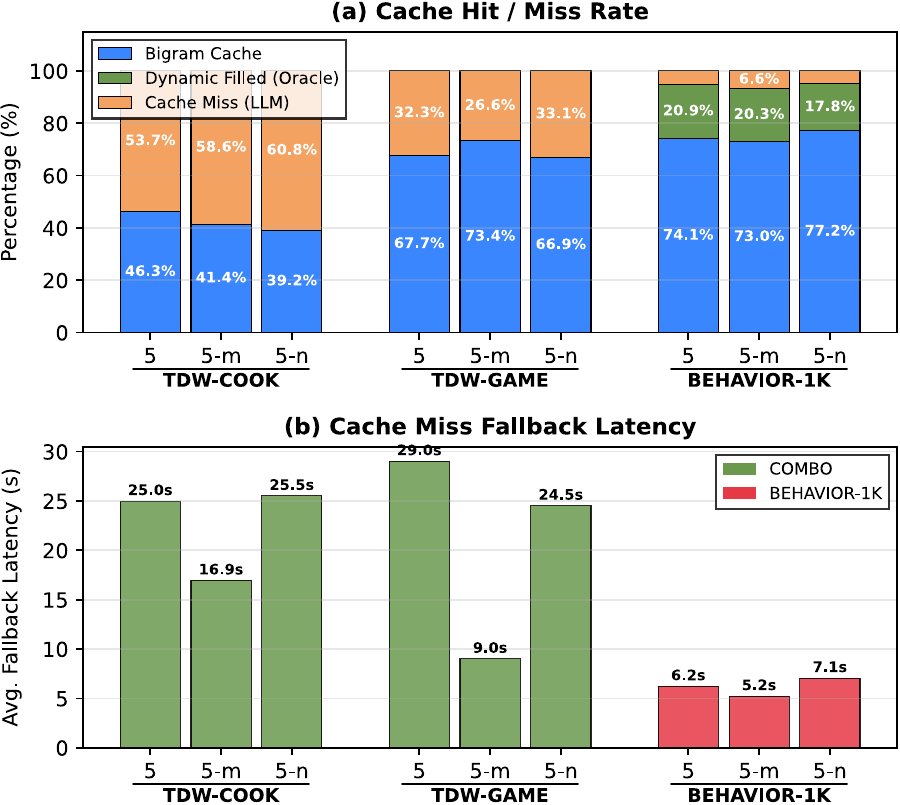}
    \caption{Cache performance across benchmarks and model scales. (a) Hit/miss rate breakdown. (b) Fallback latency on cache misses.}

    \label{fig:cache_miss_tail}
\end{figure}

\subsection{Cache Hit/Miss Rate and Fallback Latency}
\label{sec:hitmiss}

\paragraph{Results and Analysis.}
Figure~\ref{fig:cache_miss_tail}(a) shows that the bigram cache achieves high hit rates in structured environments, reaching over 66\% on TDW-GAME and at least 73\% on BEHAVIOR-1K. On TDW-COOK, hit rates drop to 39--46\% due to greater plan diversity, with remaining accesses falling back to LLM queries. Figure~\ref{fig:cache_miss_tail}(b) reports fallback latency on cache misses. TDW-based tasks incur 9--29\,s per fallback due to vision-language model overhead, while BEHAVIOR-1K maintains low latency at 5.2--7.1\,s. These results show that \name's efficiency gains are most pronounced in environments with high plan regularity, and that minimizing cache misses is critical in latency-sensitive deployments.

\subsection{Cache Size and Memory Dynamics}
\label{sec:memory}

We next check whether the cache causes meaningful memory overhead or unbounded growth during long episodes.

\begin{table}[t]
\centering
\footnotesize
\setlength{\tabcolsep}{4pt}
\renewcommand{\arraystretch}{1.05}
\caption{Cache statistics per agent. $N$ denotes the number of stored transitions, $M$ the number of metadata fields used for filtering, and Size the memory footprint $N \times (4 + \sum_{i=1}^{M} s_i)$ bytes, where $s_i = 4$ for numeric fields and $s_i = 1$ for binary fields.}
\label{tab:cache_overhead}
{
\begin{tabular*}{\linewidth}{@{\extracolsep{\fill}}lccc}
\toprule
Task & $N$ & $M$ & Size (KB) \\
\hline
CoELA@TDW-MAT & 35 & 7 & 1.0 \\
COMBO@TDW-COOK & 37 & 3 & 0.3 \\
COMBO@TDW-GAME & 35 & 1 & 0.2 \\
COHERENT@BHV-1K (Robot Arm) & 5 & 3 & 0.1 \\
COHERENT@BHV-1K (Quadrotor) & 9 & 2 & 0.1 \\
COHERENT@BHV-1K (Robot Dog) & 15 & 2 & 0.1 \\
\bottomrule
\end{tabular*}
}
\end{table}

\begin{table*}[t]
\centering
\footnotesize
\setlength{\tabcolsep}{5pt}
\renewcommand{\arraystretch}{1.05}
\caption{Growth dynamics. Average number of stored transitions $N$ across episode steps.}
\label{tab:growth_dynamics}
{
\begin{tabular}{lccccccccc}
\toprule
Model & $T{=}0$ & 500 & 1000 & 1500 & 2000 & 3000 & 4000 & 5000 & 6000 \\
\hline
GPT-5 & 0 & 3.5 & 6.0 & 7.0 & 8.1 & 10.1 & 11.7 & 12.7 & 13.7 \\
GPT-5-mini & 0 & 3.4 & 5.5 & 7.6 & 9.2 & 11.1 & 12.6 & 13.4 & 13.6 \\
GPT-5-nano & 0 & 3.4 & 5.1 & 6.5 & 7.7 & 9.8 & 11.1 & 12.0 & 12.8 \\
\bottomrule
\end{tabular}}
\end{table*}

\paragraph{Results and Analysis.} The cache footprint remains extremely small across tasks, ranging from 0.1~KB to 1.0~KB per agent (Table~\ref{tab:cache_overhead}). Transition counts (Table~\ref{tab:growth_dynamics}) grow quickly during the early phase of each run, then slow noticeably after roughly 1{,}500 steps. This pattern reflects the cache rapidly absorbing recurring transitions and then refining existing entries rather than continuously allocating new ones. Together, these results indicate that \name imposes negligible memory overhead while avoiding unbounded cache growth.

\subsection{Ablation Study}
\label{sec:ablation}
\begin{figure}[t]
  \centering
\includegraphics[width=\linewidth]{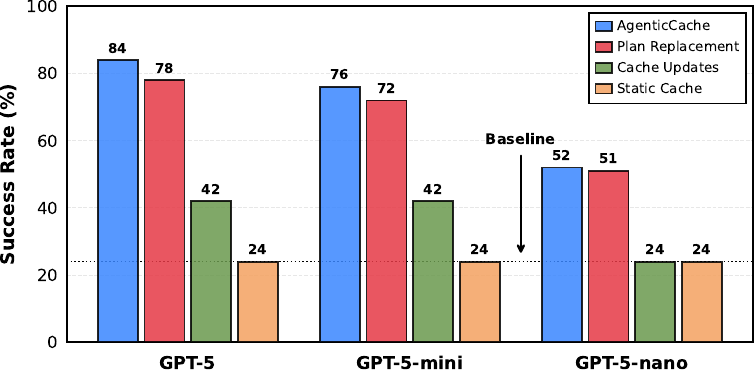}
\caption{Ablation of \name components on TDW-MAT, comparing static cache, cache updates only, plan replacement only, and the full system.}
\label{fig:ablation}
\end{figure}

\paragraph{Experimental Setup.}
We ablate \name's two main components, asynchronous cache updates and plan replacement. Four variants are evaluated: (1) a static cache without either mechanism, (2) cache updates only, (3) plan replacement only, and (4) the full system combining both.

\paragraph{Results and Analysis.}
As shown in Figure~\ref{fig:ablation}, enabling cache updates alone improves task success by 12\%, reflecting better adaptation to dynamic observations. Plan replacement further contributes a 35\% gain on average by correcting mispredicted actions on the fly. When both mechanisms are combined, the system achieves an average success rate of 70.7\%, outperforming the static baseline at 24\%. 
These results confirm that cache updates and plan replacement act synergistically, addressing complementary failure modes.

\subsection{Cache Validity Analysis}
\label{sec:validity}

\paragraph{Experimental Setup.}
To evaluate the reliability of cached plans over time, we measure the Plan Execution Accuracy. At each frame, an action is judged correct if it matches the plan that GPT-5 would have selected in the same state. The metric is computed as the difference between the cumulative correct and wrong plan frames, normalized by the current frame number. We overlay the final task success rates for \name (Ours+) and the synchronous baseline (BL) on the right axis for reference. Using GPT-5 as the reference planner, since it is the strongest model in our evaluation and best approximates ground-truth plans, we compare three configurations: \name with GPT-5, GPT-5-mini, and GPT-5-nano, all with dynamic cache updates.

\begin{figure}[t]
  \centering
\includegraphics[width=\linewidth]{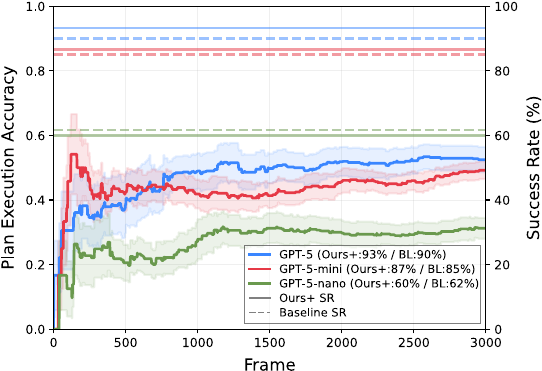}
\caption{Plan execution accuracy for \name with GPT-5, GPT-5-mini, and GPT-5-nano over time.}
\label{fig:cachehit_miss}
\end{figure}

\paragraph{Results and Analysis.}
As shown in Figure~\ref{fig:cachehit_miss}, all three dynamic \name variants show steadily improving plan execution accuracy over time. GPT-5 (blue) achieves the highest accuracy, rising to approximately 0.52 by frame 3000, followed by GPT-5-mini (red) at approximately 0.49 and GPT-5-nano (green) at approximately 0.31. This ordering aligns with model capability: stronger models produce higher-quality cache updates that better approximate oracle planning, consistent with their stronger base reasoning.

Interestingly, GPT-5-mini briefly surpasses GPT-5 in the early frames (around frame 100--500). This is because GPT-5-mini has lower inference latency, enabling shorter cache update cycles; as a result, GPT-5-mini accumulates more frequent cache refreshes in the initial phase, temporarily outpacing the slower but higher-quality updates of GPT-5. As more frames elapse, GPT-5's superior plan quality compensates for its longer update interval, and it ultimately achieves the highest accuracy.

Notably, the success rate annotations on the right axis confirm that \name (Ours+) matches or slightly exceeds the synchronous baseline (BL) for GPT-5 (93\% vs.\ 90\%) and GPT-5-mini (87\% vs.\ 85\%), while GPT-5-nano shows comparable performance (60\% vs.\ 62\%). These results demonstrate that even moderate plan execution accuracy is sufficient to maintain competitive task success, validating that \name's cache-driven planning remains reliable even when exact plan matches are imperfect.

\subsection{Discussion}
\label{sec:eval-discussion}

\paragraph{Memory Lifespan and Short-Term Recall.}
In computational systems, not all memories are worth retaining for long durations~\cite{li2025gainsight}. Accessing large-scale memory structures, such as GPU high-bandwidth memory (HBM) or external vector databases, incurs significant latency and energy costs compared to local on-chip storage. The same principle applies to embodied agents: most contextual information in embodied interaction is short-lived and quickly overwritten by new sensory input, making it inefficient to rely solely on global memory retrieval. \name provides an architectural analogue to short-term memory, storing transient yet behaviorally relevant plan transitions directly within the agent. By keeping frequently reused plans close to the execution loop and updating them asynchronously, \name achieves fast recall, low overhead, and responsiveness without the cost of global memory retrieval.

\paragraph{Toward Lifelong Adaptation.}
\name begins with an empty cache that depends on LLM guidance, but through repeated interaction it fills and refines this cache until familiar transitions can be recalled instantly, transforming episodic experience into reusable procedural memory. This mirrors the shift from deliberate reasoning to intuitive action, enabling stable yet adaptive behavior across long deployments of agentic systems with less LLM reliance.

\section{Limitations \& Future Work}
\label{sec:limitations}

\paragraph{Failure Case: Multi-Agent Coordination and Resource Contention.}
In multi-agent environments, cache reuse can fail at coordination points. For example, one agent may correctly follow a locally frequent transition while another deviates due to different observations or update timing, resulting in rendezvous mismatches such as duplicated effort or missed handoffs. A related issue arises around shared resources (e.g., cutting boards, interaction zones): a cached transition valid in isolation may cause contention or brief deadlock cycles when another agent simultaneously claims the same resource. The updater can often correct these behaviors, but the correction may arrive after several steps.

\paragraph{Scope of Plan Locality.}
The benchmarks evaluated in this work primarily involve structured manipulation and transport tasks, where plan transitions exhibit strong regularity. In more open-ended settings, such as free-form exploration or creative problem-solving, plan locality may be weaker, potentially reducing cache hit rates and limiting the benefits of \name. Evaluating the framework in less structured environments remains an important direction.

\paragraph{Future Directions.}
Several directions can address the above limitations. First, extending the cache to higher-order transition representations, such as 3-gram indexing or hierarchical subroutine fragments, could better capture delayed dependencies that 2-gram locality misses. Second, priority-based coordination protocols, including resource reservations and lightweight conflict-resolution rules, could align local cache efficiency with global multi-agent consistency. Third, the cache itself could be made more adaptive, for example by learning its scoring function or when to defer to the LLM, based on execution feedback. A broader open question is whether similar cache designs can benefit less structured domains, where plan locality is weaker.

\section{Related Work}
\label{sec:related}

\paragraph{Caching and Memoization in LLM Systems.}
Recent work has explored caching mechanisms for LLMs. MemGPT~\cite{packer2023memgpt} maintains conversational memory across sessions, while Agentic Plan Caching~\cite{zhang2025cost} stores reusable plan templates to reduce inference costs. At the semantic level, systems such as GPTCache~\cite{bang2023gptcache} and Semantic Cache~\cite{jonsson2006performance,dar1996semantic} store query--response pairs for exact or similar queries. These approaches, however, focus on token-level or response-level caching rather than plan-level patterns. 
\name uniquely exploits temporal dependencies between sequential decisions in embodied tasks, where recurring plan transitions create a caching opportunity that prior work does not target.

\paragraph{Predictive Prefetching and Speculative Execution.}
Our work draws inspiration from computer architecture. Branch prediction~\cite{smith1998study,yeh1993comparison,seznec2007256,villon2023conditional,10.1145/2155620.2155635} and speculative execution~\cite{gabbay1996speculative,10.1145/3399742,10.1145/290409.290411,moshovos2002microarchitectural} continue computation based on predicted outcomes, analogous to how \name executes predicted plans while awaiting LLM responses. More recently, SpecInfer~\cite{miao2024specinfer} and Medusa~\cite{cai2024medusa} apply speculative decoding to accelerate LLM inference. \name extends these ideas to agent planning, treating plans as coarse-grained ``branches'' that can be predicted and speculatively executed.

\paragraph{Learning from Demonstrations and Trajectory Replay.}
Imitation learning methods~\cite{abbeel2004apprenticeship,ross2011reduction} and trajectory optimization techniques~\cite{todorov2005generalized,kalakrishnan2011stomp} learn policies from expert demonstrations. Similarly, ALFRED~\cite{ALFRED20} and TEACh~\cite{padmakumar2022teach} collect human demonstrations for embodied instruction following. While these approaches use offline data to train policies, \name leverages demonstrations differently: it extracts frequent transition patterns for cache initialization, enabling immediate deployment without retraining. In our evaluation, this requires only 1--4 training episodes per benchmark, compared with thousands typical for policy learning.

\section{Conclusion}
\label{sec:conclusion}

We present \name, a cache-driven planning framework that reuses frequent plan transitions to avoid per-step LLM calls in embodied AI agents. Each agent queries a runtime cache of plan transitions at every step, while a background Cache Updater asynchronously validates and refines cached entries with the LLM. This design exploits the strong plan locality we observed in embodied tasks, keeping the agent responsive without blocking on LLM inference.

Across four multi-agent embodied benchmarks and three model scales, \name improves task success rate by 22\% on average, reduces simulation latency by 65\%, and lowers token usage by 50\%. These results indicate cache-based plan reuse as a practical lever for low-latency, low-cost embodied agents. Similar cache designs may benefit other agent settings where short-horizon decisions follow predictable patterns, and we leave this to future work.

\section*{Acknowledgements}
This research was supported in part with gifts from Apple and Google. We thank Yeonjae Kim (SNU) and Yeonhong Park (Meta) for their insightful feedback on our work.




\bibliography{references}
\bibliographystyle{mlsys2026}

\appendix
\section{Artifact Appendix}
\label{sec:artifact}

\subsection{Abstract}

This artifact contains the source code and evaluation scripts for \textbf{\name}, a cache-driven asynchronous planning framework for LLM-based embodied agents. The artifact reproduces the main experimental results across four multi-agent embodied benchmarks: COHERENT (BEHAVIOR-1K), CoELA (TDW-MAT), and COMBO (TDW-COOK, TDW-GAME). Each benchmark is provided as a Git submodule with four branches (\texttt{agenticcache}, \texttt{baseline}, \texttt{parallel}, \texttt{speculative}) corresponding to the methods compared in the paper. COMBO additionally includes a \texttt{training-code} branch for reproducing the vision diffusion model.

\emph{Note:} Due to the non-deterministic nature of LLM inference, exact numerical reproduction is not guaranteed. We provide our original running logs in the GitHub repository, and reproduced results are expected to be consistent with the trends and magnitudes reported in the paper.

\subsection{Artifact Check-List (Meta-Information)}

\begin{itemize}
  \item \textbf{Algorithm:} \name (cache-driven asynchronous LLM planning)
  \item \textbf{Program:} Python
  \item \textbf{Data set:} COHERENT (BEHAVIOR-1K), CoELA (TDW-MAT), COMBO (TDW-COOK, TDW-GAME)
  \item \textbf{Run-time environment:} Linux, conda, CUDA, Python 3.10+
  \item \textbf{Hardware:} GPU required (NVIDIA A100 or equivalent recommended); TDW simulator requires X11 display
  \item \textbf{Execution:} Automated via shell scripts per benchmark
  \item \textbf{Metrics:} Task success rate, latency, token usage, cost
  \item \textbf{Output:} JSON result logs per episode
  \item \textbf{Experiments:} Table~\ref{tab:planning_results}, Figure~\ref{fig:EmbodiedAIAgent}, Figure~\ref{fig:ActionLocality}.
  \item \textbf{How much disk space required (approximately)?:} $\sim$200\,MB for code and datasets; $\sim$1.5\,GB additional if reproducing COMBO training (model checkpoint)
  \item \textbf{How much time is needed to prepare workflow (approximately)?:} $\sim$2\,days (COMBO model training on 2$\times$H100); evaluation setup $<$1\,hour per benchmark
  \item \textbf{How much time is needed to complete experiments (approximately)?:} $\sim$10\,days for all benchmarks (3 submodules $\times$ 4 methods)
  \item \textbf{Publicly available?:} Yes
  \item \textbf{Workflow framework used?:} conda, shell scripts
\end{itemize}

\subsection{Description}

\subsubsection{How Delivered}

The artifact is delivered as a GitHub repository with three Git submodules:
\begin{itemize}
  \item \texttt{MLSys26\_AgenticCache-COHERENT} -- COHERENT (BEHAVIOR-1K) benchmark evaluation
  \item \texttt{MLSys26\_AgenticCache-CoELA} -- CoELA (TDW-MAT) benchmark evaluation
  \item \texttt{MLSys26\_AgenticCache-COMBO} -- COMBO (TDW-COOK, TDW-GAME) benchmark evaluation
\end{itemize}
Each submodule contains branches \texttt{agenticcache}, \texttt{baseline}, \texttt{parallel}, and \texttt{speculative}. COMBO additionally has a \texttt{training-code} branch.

Clone with: \texttt{git clone --recursive} \url{https://github.com/hojoonleokim/MLSys26_AgenticCache.git}

\subsubsection{Hardware Dependencies}

NVIDIA GPU with CUDA support and $\geq$24\,GB VRAM is required.
A display server (X11) is needed for the TDW simulator (can use virtual display via \texttt{Xvfb}).
Tested on:
\begin{itemize}
  \item \textbf{Evaluation:} AMD Ryzen 9 7950X (16-core), 128\,GB RAM, NVIDIA GeForce RTX 4090 (24\,GB), Ubuntu 22.04 LTS.
  \item \textbf{Training (COMBO):} AMD EPYC 9454 (48-core), 2.2\,TB RAM, 2$\times$ NVIDIA H100 PCIe (80\,GB), Ubuntu 22.04 LTS.
\end{itemize}

\subsubsection{Software Dependencies}

\begin{itemize}
  \item Linux (tested on Ubuntu)
  \item conda (Anaconda or Miniconda)
  \item CUDA toolkit
  \item OpenAI API key (for GPT-5 inference)
  \item TDW simulator (CoELA and COMBO)
\end{itemize}
Per-benchmark conda environments are defined in each submodule (\texttt{environment.yml}).

\subsubsection{Datasets}

All evaluation datasets are bundled within the submodules. No external download is required for evaluation. For COMBO training, data is generated via the TDW simulator as part of the training pipeline (see \texttt{training-code} branch).

\subsection{Installation}

\begin{enumerate}
  \item Clone the repository with submodules:

\texttt{git clone --recursive} \url{https://github.com/hojoonleokim/MLSys26_AgenticCache.git}
\begin{verbatim}
cd MLSys26_AgenticCache
\end{verbatim}

  \item Create conda environments from the \texttt{environment.yml} in each submodule (on the \texttt{baseline} branch):
\begin{verbatim}
P=MLSys26_AgenticCache
conda env create \
  -f $P-COHERENT/environment.yml
conda env create \
  -f $P-CoELA/environment.yml
conda env create \
  -f $P-COMBO/environment.yml
\end{verbatim}
  This creates conda environments named \texttt{coherent}, \texttt{tdw}, and \texttt{combo}, respectively.

  \item Set the \texttt{OPENAI\_API\_KEY} environment variable for GPT-5 access.

  \item (CoELA \& COMBO only) Set up the X server for TDW. Kill any existing display server processes, then start Xorg:
\begin{verbatim}
# Kill existing Xorg / gnome-shell
sudo kill -9 <PID_of_Xorg>
sudo kill -9 <PID_of_gnome-shell>

# Start X server on display :1
sudo nohup Xorg :1 \
  -config /etc/X11/xorg-1.conf &
\end{verbatim}
  See the \href{https://github.com/threedworld-mit/tdw/blob/master/Documentation/lessons/setup/server.md}{TDW server setup guide} for generating \texttt{xorg.conf} files.

  \item (COMBO only) To reproduce the vision model from scratch, switch to the \texttt{training-code} branch and run the training pipeline:
\begin{verbatim}
cd MLSys26_AgenticCache-COMBO
git checkout training-code
cd AVDC/flowdiffusion
bash train_all.sh
\end{verbatim}
  The pipeline consists of four steps: (1)~conda env setup, (2)~training data generation via TDW (requires \texttt{DISPLAY=:1}), (3)~text embedding preprocessing with T5-XXL, and (4)~inpainting diffusion model training (100K steps). The final checkpoint \texttt{modl-100.pt} is used by all evaluation branches.
\end{enumerate}

\subsection{Experiment Workflow}

Automated scripts in \texttt{scripts/} iterate over all four branches (\texttt{baseline}, \texttt{agenticcache}, \texttt{parallel}, \texttt{speculative}), check out each branch, and run the experiments:

\begin{verbatim}
# COHERENT (no Xorg needed)
./scripts/run_coherent.sh

# CoELA (requires Xorg on :1)
./scripts/run_coela.sh

# COMBO (requires Xorg on :1)
./scripts/run_combo.sh
\end{verbatim}

We prefill the cache with the following episodes, which are held out from the evaluation set:
\begin{itemize}
  \item \textbf{COHERENT (BEHAVIOR-1K):} \texttt{env0/task\_15}, \texttt{env1/task\_10}, \texttt{env2/task\_11}, \texttt{env3/task\_16}
  \item \textbf{CoELA (TDW-MAT):} test episodes 1--4
  \item \textbf{COMBO (TDW-COOK, TDW-GAME):} cook episodes 0--1, game episode 0
\end{itemize}

Each script runs three model variants (GPT-5, GPT-5-mini, GPT-5-nano) sequentially. Results are saved as JSON logs under each benchmark's \texttt{results/} directory.

\noindent\textbf{Estimated runtime per branch (single GPU):}
\begin{itemize}
  \item COHERENT (BEHAVIOR-1K): $\sim$2\,hours (graph-only, no simulator)
  \item CoELA (TDW-MAT): $\sim$12\,hours
  \item COMBO (TDW-COOK + TDW-GAME): $\sim$8\,hours
\end{itemize}

\subsection{Evaluation and Expected Result}

The expected results correspond to the evaluation results reported in the main paper.
The raw result logs used to produce these figures and tables are stored in the \texttt{results/} directory of the main Git repository.

Due to the stochastic nature of LLM inference, exact numerical results will vary across runs. Reviewers should verify that:
\begin{enumerate}
  \item \name (\texttt{agenticcache} branch) matches or outperforms the baseline in task success rate.
  \item \name reduces simulation latency compared to the synchronous baseline.
  \item \name reduces total token usage.
  \item The parallel and speculative variants show distinct trade-offs compared to the baseline and \name.
\end{enumerate}

\subsection{Experiment Customization}

Reviewers may customize the evaluation as follows:
\begin{itemize}
  \item \textbf{Run a single branch:} Instead of the automated scripts, manually check out a specific branch and run the per-benchmark script (e.g., \texttt{scripts/test\_LMs-gpt-5.sh} for CoELA).
  \item \textbf{Change the LLM:} Edit the \texttt{MODELS} array in each benchmark's internal script (e.g., \texttt{scripts/run\_all.sh} for COHERENT, \texttt{scripts/test\_LMs-gpt-5.sh} for CoELA, \texttt{scripts/run\_gpt5\_all.sh} for COMBO).
  \item \textbf{Adjust episode scope:} The cache episodes and evaluation episodes are defined at the top of each script and can be modified.
\end{itemize}

\subsection{Notes}

\begin{itemize}
  \item CoELA (TDW-MAT) and COMBO (TDW-COOK, TDW-GAME) require an active X server (\texttt{DISPLAY=:1}) for the TDW simulator. COHERENT (BEHAVIOR-1K) is text-only and does not require a display.
  \item The COMBO \texttt{training-code} branch is provided for full reproducibility of the vision diffusion model but is not required if using the provided checkpoint.
\end{itemize}

\subsection{Artifact Review References}

Submission, reviewing, and badging methodology:

\begin{itemize}
  \item \url{http://cTuning.org/ae/submission-20190109.html}
  \item \url{http://cTuning.org/ae/reviewing-20190109.html}
  \item \url{https://www.acm.org/publications/policies/artifact-review-badging}
\end{itemize}

\section{Prompt Templates}
\label{sec:appendix:prompt}

This appendix provides the prompt templates used in our experiments. 
Figure~\ref{fig:TDW-MAT}-\ref{fig:BEHAVIOR-1K} show example templates for the four embodied tasks: TDW-MAT, TDW-COOK, TDW-GAME, and BEHAVIOR-1K. 

Each template specifies the agent's role, operational constraints, current observation, oracle instruction, action history, and the set of available actions. 
The LLM is instructed to select exactly one valid action per step following a strict output format to ensure consistent reasoning and reproducible evaluation.

\begin{figure*}[t] 
\centering
\begin{tcolorbox}[title=Prompt for TDW-MAT,
  colback=gray!5, colframe=gray!50,
  boxrule=0.8pt, arc=4pt, outer arc=4pt, 
  width=\textwidth]  

\begin{lstlisting}[basicstyle=\ttfamily\small, breaklines=true]
I'm $AGENT_NAME$. My teammate $OPPO_NAME$ and I are collaborating to transport as many target objects as possible to the bed within 3000 steps. 
We can use containers early for efficiency, but as objects or containers become scarce, direct hand transport may be faster. Reason adaptively based on our current progress.

Rules and Abilities:
- I can hold up to two items (objects or containers).
- Each container holds up to three objects and is lost once delivered to the bed.
- Maximum capacity: 2 objects without containers, 4 objects with a container (3+1).
- Moving between rooms or transporting to the bed is costly - use sparingly.
- Objects are denoted as <name> (id), e.g., <apple> (712).
- `go grasp target` and `go grasp container` list nearby items first, then others.
- Coordinate with my teammate to avoid redundant efforts.
- **Exploration rule:** Alice mainly explores even-numbered rooms, Bob odd-numbered ones. This is a guideline, not a strict rule - assist if it's more efficient.

Action Decision Rules:
- Actions with suffix `canceled` were canceled before completion by your new decision.
- Actions with suffix `unreached` failed to reach the target destination after many attempts.
- The last action is the **current ongoing action**.
  - You may cancel it if a better one exists, but it incurs **opportunity cost** (e.g., when few steps remain, both agents go to the same room, or I'm wasting time).
  - If the current action has been running too long without progress, I may be **stuck** - switch to a better one.
  - If the current action is still effective, it's fine to **keep the current action**.
- **Efficiency tip:** Choose "transport" directly instead of "keep current action: go to Bedroom" - transport completes delivery in one action, while going to bedroom then transporting requires two actions.
- Select exactly one **best available action** that efficiently contributes to the goal.

Context:
Goal: $GOAL$
MY $PROGRESS$

Dialogue history:
$DIALOGUE_HISTORY$

Previous actions:
$ACTION_HISTORY$

Completed objects: $COMPLETED_OBJECTS$/10

Available actions:
$AVAILABLE_ACTIONS$

Think step by step to select the best action. After your reasoning, try to provide your final answer on a new line in this exact format:
ANSWER: [letter]

For example:
ANSWER: A
\end{lstlisting}

\end{tcolorbox}
\caption{Prompt for TDW-MAT.}
\label{fig:TDW-MAT}
\end{figure*}

\begin{figure*}[!p] 
\centering
\begin{tcolorbox}[title=Prompt for TDW-COOK,
  colback=gray!5, colframe=gray!50,
  boxrule=0.8pt, arc=4pt, outer arc=4pt, 
  width=\textwidth]  

\begin{lstlisting}[basicstyle=\ttfamily\scriptsize, breaklines=true]
Two agents, Alice and Bob, are cooperating at a kitchen counter. Each agent can operate only along one counter edge. The goal is to make a burger and a sandwich.

Some food items must be cut on the cutting board to obtain the required ingredients.

You are agent {agent_name}, near the {agent_pos} edge of the counter, making {recipe_strs[agent_id].split(":")[0]}.  
You can only pick/place items **along your edge** and cut items **on the cutting board**.

There are exactly as many food items as needed-no more, no less. All items must be used. The task ends when both agents finish their recipes.

### Shared Cutting Board Cooperation
- The **cutting board** is the **only shared region** (capacity: 1 item).  
  Use it to **cut items** or **pass items** the other agent needs.
- Each agent also has a **private region** (4 slots) to **temporarily store** items or prevent shared cutting board congestion.
- If an item **is not part of your recipe**, place it on the cutting board **only when the other agent can likely pick it soon**; otherwise keep it in your private region.
- If the cutting board holds an item **you need**, take it immediately.
- The environment automatically ensures that:
  - "place onto plate" appears **only when the ingredient matches the next recipe order**, and  
  - "pick up" appears **only for reachable items**.
- Avoid blocking the cutting board with unnecessary items.  
  Each step should either (1) advance your recipe, or (2) help the other agent progress.

### Decision
Given the current visual scene shown in the image, select **exactly one** action from the list below.
All listed actions are **guaranteed valid** - the environment only includes actions that are logically possible.

**Selection Rules (in order):**
1. Prefer actions that immediately advance your own recipe.
   - You can identify the needed ingredient by checking your recipe, action history and dish.
2. Otherwise, help the other agent without blocking the cutting board.
   - You can identify the needed ingredient by checking the other agent's recipe and dish.
3. Use your private region to temporarily store items or to prevent congestion on the shared cutting board.

### Progress
You've taken **{steps_taken-1}/60** steps.  
Steps remaining: **{60 - (steps_taken-1)}**.

Recipe (stack in order):
- Yours: {recipe_strs[agent_id]}
- Other agent: {recipe_strs[1 - agent_id]}

Action History (excluding WAIT):
#ACTION_HISTORY#

Possible Actions:  
#POSSIBLE_ACTIONS#

Output (strictly one line, no reasoning):  
Next action: <one of the listed actions>
\end{lstlisting}
\end{tcolorbox}
\caption{Prompt for TDW-COOK.}
\label{fig:TDW-COOK}
\end{figure*}

\begin{figure*}[t] 
\centering
\begin{tcolorbox}[title=Prompt for TDW-GAME,
  colback=gray!5, colframe=gray!50,
  boxrule=0.8pt, arc=4pt, outer arc=4pt, 
  width=\textwidth]  

\begin{lstlisting}[basicstyle=\ttfamily\small, breaklines=true, mathescape=true]
Four agents {agents_name_str} are cooperating around a square table to solve a puzzle game.  
Each agent can operate only within the region along one edge (north, east, south, west).  
The goal is to place all puzzle pieces into their correct positions inside the puzzle box.

You are agent {agent_name}, near the {agent_pos} edge.  
You can only interact with pieces **along your edge**, including both **borders** of your reachable region.

### Shared-Border Cooperation
- Each table border is a **shared region** between two adjacent agents.  
  It is used **to pass pieces** between them (capacity: 1 piece per border).
- Each agent also has **two private regions** for temporarily holding or prevent shared border congestion.
- If you hold a piece **not needed for your puzzle**, place it on a shared border **only when the adjacent agent can likely pick it soon**; otherwise keep it in your private region.
- If a shared border has a piece, pick it up to check if it fits your puzzle; otherwise, pass it to others.
- The environment automatically ensures that:
  - "place into puzzle box" appears only when the held piece **belongs to your puzzle**, and  
  - "pick up" appears only for **reachable pieces**.

### Action Selection
Given the current visual scene shown in the image, select **exactly one** action from the list below.
All listed actions are **guaranteed valid** - the environment filters out unreachable or invalid actions.

**Selection Rules (in order):**
1. Prefer placing your correct piece into the puzzle box.
   - You can decide which piece to place by checking your puzzle box.
2. If (1) is not possible, pass the piece via a shared border.
   - You can choose which border to use by checking the other agents' puzzle boxes.
3. Use private regions to temporarily store pieces or to prevent congestion of the shared borders.

### Progress
You've taken **{steps_taken-1}/60** steps.  
Steps remaining: **{60 - (steps_taken-1)}**.

Action History (excluding WAIT):  
#ACTION_HISTORY#

Possible Actions:  
#POSSIBLE_ACTIONS#

Output (strictly one line, no reasoning):  
Next action: <one of the listed actions>
\end{lstlisting}
\end{tcolorbox}
\caption{Prompt for TDW-GAME.}
\label{fig:TDW-GAME}
\end{figure*}

\begin{figure*}[t] 
\centering
\begin{tcolorbox}[title=Prompt for BEHAVIOR-1K,
  colback=gray!5, colframe=gray!50,
  boxrule=0.8pt, arc=4pt, outer arc=4pt, 
  width=\textwidth]  

\begin{lstlisting}[basicstyle=\ttfamily\small, breaklines=true, mathescape=true]
You are a **robot arm** in a cooperative multi-agent household task environment.  
You are fixed on a table or platform and can operate objects **only on that surface**.  
You can pick and place objects on the table, open or close containers on the same table,  
and interact with the **basket of the quadrotor** if the quadrotor lands on your table.  
Objects on other tables or in other locations are out of reach.  
If the quadrotor lands on a different table, you cannot interact with its basket.  
When holding something, you cannot open or close doors or containers.  
Avoid redundant actions that repeat an already completed result.

---

### Current Context
- **Observation:**  
  #OBSERVATION#
- **Oracle Instruction:**  
  #INSTRUCTION#
- **Action History:**  
  #PLANHISTORY#
- **Available Actions:**  
  #ACTIONLIST#

---

### Your Task
Choose **exactly one** best action from the available actions that will most effectively advance the instruction.
Look at Observation to understand the current state of the environment.
If none of the actions can progress the task, explain briefly why (for example: out of reach, limited ability, missing prerequisite, or goal already achieved).
Use oracle instruction to guide your actions. Some actions might already be achieved refer to action history.
If you don't have a clue to achieve the oracle instruction, you can just wait.
All object is denoted as <class name>(id). It is exclusive and unique.

---

### Output Format (must follow exactly one)
- If you can proceed:  
  ACTION: <copy exactly one action string from the list above>

- If you cannot proceed:  
  SORRY, I CANNOT: <one-sentence reason>

---

### Output Rules
1. Copy the action string **verbatim** from the list if you can act.  
2. Do **not** add extra words, numbers, or multi-line explanations.  
3. Do **not** invent or rephrase actions.  
4. Always think step by step, but output only one final line in the required format.
5. If you finished the oracle instruction, output [wait].
\end{lstlisting}
\end{tcolorbox}
\caption{Prompt for BEHAVIOR-1K.}
\label{fig:BEHAVIOR-1K}
\end{figure*}


\end{document}